\documentclass[letterpaper]{article} 
\usepackage{aaai25}  
\usepackage{times}  
\usepackage{helvet}  
\usepackage{courier}  
\usepackage[hyphens]{url}  
\usepackage{graphicx} 
\usepackage{natbib}  
\usepackage{caption} 
\usepackage{algorithm}
\usepackage{algorithmic}
\usepackage{graphicx}
\usepackage{amsmath} 
\usepackage{float}
\usepackage[nohyperref]{xurl}
\usepackage{amssymb}
\usepackage{enumitem}
\usepackage{multirow}

\usepackage{newfloat}
\usepackage{listings}
\DeclareCaptionStyle{ruled}{labelfont=normalfont,labelsep=colon,strut=off} 
\floatstyle{ruled}
\newfloat{listing}{tb}{lst}{}
\floatname{listing}{Listing}
\title{Enhancing Vietnamese VQA through Curriculum Learning on Raw and Augmented Text Representations}

\author {
    Anh-Khoi Nguyen\textsuperscript{\rm 1,\rm 2}\thanks{These authors contributed equally.},
    Yen-Linh Vu\textsuperscript{\rm 1}\footnotemark[1],
    Dinh-Thang Duong\textsuperscript{\rm 1}\footnotemark[1],
    Nguyen-Thuan Duong\textsuperscript{\rm 1},
    Thanh-Huy Nguyen\textsuperscript{\rm 1,3},
    Quang-Vinh Dinh\textsuperscript{\rm 1}\thanks{Corresponding author: vinh.dinhquang@aivietnam.edu.vn}
}

\affiliations {
    \textsuperscript{\rm 1}AI VIETNAM Lab, Vietnam\\
    \textsuperscript{\rm 2}Ton Duc Thang University, Ho Chi Minh City, Vietnam\\
    \textsuperscript{\rm 3}Université de Bourgogne, Dijon, France
}

\begin{document}

\maketitle

\begin{abstract}
Visual Question Answering (VQA) is a multimodal task requiring reasoning across textual and visual inputs, which becomes particularly challenging in low-resource languages like Vietnamese due to linguistic variability and the lack of high-quality datasets. Traditional methods often rely heavily on extensive annotated datasets, computationally expensive pipelines, and large pre-trained models, specifically in the domain of Vietnamese VQA, limiting their applicability in such scenarios. To address these limitations, we propose a training framework that combines a paraphrase-based feature augmentation module with a dynamic curriculum learning strategy. Explicitly, augmented samples are considered ``easy'' while raw samples are regarded as ``hard''. The framework then utilizes a mechanism that dynamically adjusts the ratio of easy to hard samples during training, progressively modifying the same dataset to increase its difficulty level. By enabling gradual adaptation to task complexity, this approach helps the Vietnamese VQA model generalize well, thus improving overall performance. Experimental results show consistent improvements on the OpenViVQA dataset and mixed outcomes on the ViVQA dataset, highlighting both the potential and challenges of our approach in advancing VQA for Vietnamese language. 
\end{abstract}

\begin{links}
    \link{Code}{https://github.com/wjnwjn59/CLAugViVQA}
\end{links}

\section{Introduction}

Visual Question Answering (VQA) is a complex task that involves processing both textual and visual information to answer natural language questions about images \cite{antol2015vqa}. While Transformer-based architectures have significantly advanced the field \cite{vaswani2017attention}, challenges such as linguistic variability \cite{tang2020semantic}, language biases \cite{li2019visualbert}, and limited generalization remain, particularly in low-resource languages like Vietnamese \cite{luong2021bartpho}.

In VQA, the objective is to predict the most plausible answer distribution given an image and a question. For Vietnamese, the task is especially complex due to unique syntactic and semantic structures, where paraphrased questions often differ significantly in form. Such variations create inconsistencies between training and testing datasets \cite{wang2022tag}, as the model may struggle to generalize from limited training examples to the diverse linguistic patterns seen during evaluation. Furthermore, VQA in Vietnamese is still a relatively new area of research and suffers from a lack of high-quality datasets, making it challenging to train models capable of handling the language's complexities. These challenges underscore the urgent need for training strategies that enhance generalization while effectively addressing linguistic diversity in resource-constrained scenarios. For instance, a training method for Vietnamese VQA models capable of leveraging external knowledge to improve performance without relying on additional annotated data is essential.

To alleviate these aforementioned problems, we introduce a novel training pipeline that leverages curriculum learning to incorporate both raw and augmented text representations. By considering augmented samples created through paraphrasing as ``easy'' and original samples as ``hard'', this approach dynamically adjusts the proportion of these samples presented to the model during training. Initially, the model is exposed to a higher number of easy samples to build foundational understanding, and as training progresses, the number of harder samples increases. This gradual adjustment allows the model to adapt effectively to the growing complexity of the task, enhancing its ability to handle linguistic variability and improving its robustness and generalization for Vietnamese VQA tasks.

Through extensive experimentation, it is demonstrated that the proposed pipeline effectively addresses the linguistic and computational challenges of low-resource settings, offering a promising framework for advancing VQA systems in underrepresented languages.

In summary, our contributions are as follows:
\begin{enumerate}
    \item We present a simple feature augmentation module to improve the representation of text features in Vietnamese VQA models. By utilizing information from paraphrased versions of questions, the linguistic complexity of the Vietnamese language is addressed, resulting in better text embeddings.
    \item Learning on a mixture of original and augmented embeddings is shown to lead to more effective convergence, thereby enhancing overall performance.
    \item Motivated by \cite{bengio2009curriculum}, our proposed training pipeline is adapted with the idea of curriculum learning, where input data with augmented embeddings are treated as easier samples and original embeddings as harder samples. Starting with a higher proportion of easy samples and gradually increasing the number of hard samples during training, noticeable improvements in the performance of baseline models are observed.
\end{enumerate}


\section{Related Works}

\subsection{VQA General Approaches}
Visual Question Answering (VQA) has evolved from traditional CNN-LSTM models \cite{antol2015vqa}, which struggled with linguistic variability, to Transformer-based architectures \cite{vaswani2017attention}, which significantly improved the integration of textual and visual features. Attention mechanisms, such as Bottom-Up and Top-Down Attention \cite{anderson2018bottom}, further enhanced reasoning capabilities by effectively aligning these modalities.

After Transformers demonstrated their success across various tasks, numerous works have applied these architectures to VQA. Models such as LXMERT \cite{tan2019lxmert}, VisualBERT \cite{li2019visualbert}, and BLIP \cite{li2022blip} have successfully utilized Transformers to process multimodal data, enabling effective performance on tasks like VQA and image captioning. With the advancements of Large Language Models (LLMs), efforts have been made to further improve vision-language integration by connecting a pre-trained vision encoder with LLMs and then fine-tuning on vision tasks, rather than training both text and image encoders from scratch. This approach leverages the pre-trained capabilities of LLMs, reducing the computational overhead while achieving superior performance. As demonstrated in \cite{team2023gemini, liu2023llava, internvl_cvpr, Qwen-VL}, these Large Vision-Language Models (LVLMs) excel in addressing vision-related tasks, including VQA, by combining sophisticated textual and visual reasoning capabilities.

Despite their impressive performance, these vision-language models share significant limitations, including large model sizes and high training resource requirements. We believe these factors make them less suitable for VQA-specific applications, particularly in low-resource settings like Vietnamese.

\subsection{Feature Augmentation}

Feature augmentation is a critical technique in Visual Question Answering (VQA) aimed at enhancing model robustness and generalization by diversifying training data. Several approaches have made significant progress in addressing linguistic variability and data scarcity. The SEADA method \cite{tang2020semantic} generates adversarial examples while maintaining semantic equivalence, improving generalization and robustness. Similarly, TAG \cite{wang2022tag} expands datasets by generating additional question-answer pairs, addressing linguistic diversity, and improving the performance of Text-VQA. These works highlight the importance of data variability in strengthening VQA models against linguistic complexity.

Augmentation techniques have also focused on improving multi-modal representation by addressing the interplay between textual and visual data. Cross-modal alignment approaches, such as those by \cite{chen2022rethinking} and \cite{mashrur2024robust}, or Multi-views approaches, such as \cite{nguyen2023towards, ngo2024dual, truong2023delving} leverage augmentation to refine the compatibility between modalities, leading to improved reasoning capabilities. However, these methods often rely on large, well-annotated datasets, limiting their applicability in low-resource languages like Vietnamese. Additionally, their strategies are designed for general multi-modal tasks, lacking specificity for the unique linguistic structures of Vietnamese.

Semi-Supervised Implicit Augmentation \cite{dodla2024semi} introduces feature variations implicitly, enriching training datasets without extensive labeled samples and reducing dependence on large datasets. However, it does not explicitly address linguistic variability or provide targeted solutions for underrepresented languages, where semantically diverse textual features are critical for effective learning.

Despite these advancements, limitations persist in addressing computational efficiency and language-specific variability. Most methods focus on either cross-modal consistency or data scarcity but fail to holistically tackle both issues in low-resource settings. These gaps highlight the need for a lightweight, language-specific augmentation strategy that can enhance multi-modal alignment and enrich linguistic diversity.



\begin{figure*}[ht]
    \centering
    \includegraphics[width=\textwidth]{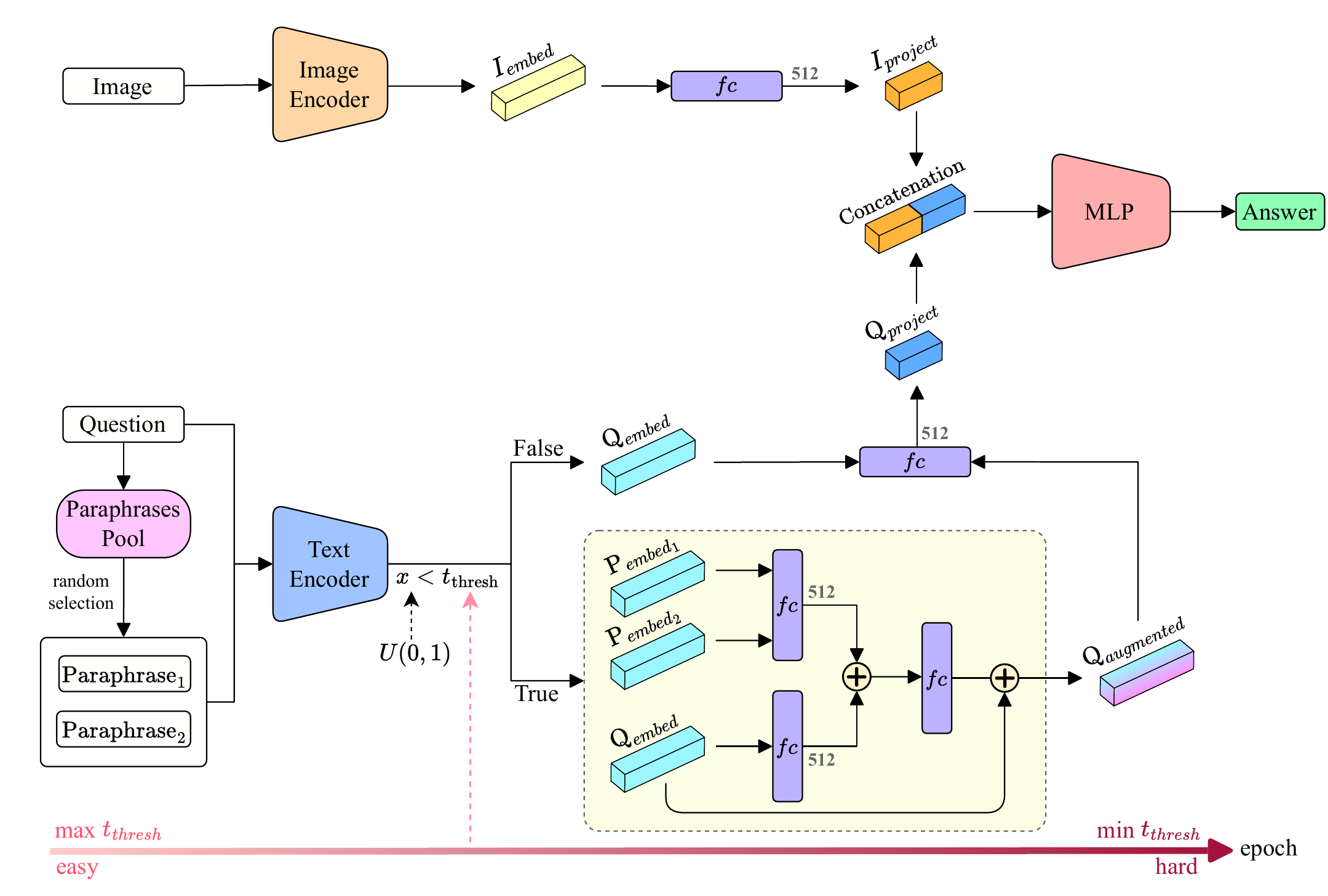}
    \caption{Overview of our proposed training pipeline utilizing curriculum learning on a mix of raw and augmented samples for a Vietnamese VQA model. The text channel produces embeddings using a text encoder, where augmented embeddings are created by combining the original question embedding with paraphrased embeddings from a paraphrase pool, governed by a threshold dynamically adjusted during training ($t_\text{thresh}$). The image channel extracts visual features using an image encoder in the standard manner. Finally, the text and image features are connected and processed through some fully-connected layers (MLP) to generate the final answer.}
    \label{fig:pipeline}
\end{figure*}

\subsection{Curriculum Learning}

Curriculum learning is a training approach that structures the learning process by progressively introducing tasks of increasing difficulty, allowing models to progress from simpler to more challenging tasks. The original concept \cite{bengio2009curriculum} demonstrated improvements in model convergence and generalization by structuring data exposure. Building on this, Dynamic Curriculum Learning (DCL) \cite{wang2019dynamic} dynamically adjusts task difficulty during training to improve performance on imbalanced datasets, showcasing its effectiveness across various domains.

Within the scope of VQA, curriculum learning has been applied to tackle challenges such as data scarcity and multi-modal reasoning complexity. For example, one approach \cite{askarian2021curriculum} leverages task progression to adapt models in low-resource scenarios, transitioning from simpler tasks to complex reasoning, significantly improving VQA performance under data constraints. Another method \cite{zheng2024improving} integrates curriculum learning with feature augmentation, enhancing alignment between textual and visual modalities while ensuring scalability in resource-constrained environments.

In spite of these improvements, current methods face limitations in addressing the unique challenges of languages like Vietnamese. Many approaches assume linguistic uniformity and overlook syntactic and semantic complexities inherent to such languages. Additionally, the computational overhead associated with dynamic task adjustment limits their practicality in low-resource contexts.

Building on this technique, the curriculum strategy for Vietnamese VQA employs a paraphrase-based augmentation module within the network. Training begins with a higher proportion of semantically diverse paraphrased questions (``easy'' samples) and gradually increases original questions (``hard'' samples) as training progresses. This seamless integration ensures robust multimodal alignment and effective generalization.

\section{Methodology}

\subsection{Problem Definition}
We consider VQA task as a multi-class classification problem. Given a dataset $\mathcal{D} = \{Q_{i}, I_{i}, a_{i}\}^{N}_{i}$ of size $N$ which each sample consists of a question $Q_{i} \in \mathcal{Q}$, an image $I_{i} \in \mathcal{I}$ and an answer $a_{i} \in \mathcal{A}$. The VQA learning objective is to be able to map $f: \mathcal{Q} \times \mathcal{I} \to \mathcal{P}(\mathcal{A})$ which presents the distribution of answer space given a question-image pair.

\subsection{Baseline model}
A dual-stream model architecture, inspired by previous works \cite{antol2015vqa, anderson2018bottom}, is adopted, where each stream can utilize any suitable encoder for text and image data. In this study, the aim is to design a lightweight and simple VQA architecture while the dual-stream model is adapted to leverage pre-trained transformer-based Vietnamese models for improved performance, as demonstrated in \cite{bartphobeit}. The text encoder (\textbf{T\_Encoder}) processes questions to produce text embeddings (\(Q\_embed\)), while the image encoder (\textbf{I\_Encoder}) extracts image embeddings (\(I\_embed\)).

After creating the text and image embeddings using their respective encoders, linear projections are applied to align these features. Specifically, the image embedding (\(I\_embed\)) is obtained by passing the input image through the image encoder, followed by a projection matrix \(W^{I}\). Similarly, the question embedding (\(Q\_embed\)) is generated by processing the input question with the text encoder and applying a projection matrix \(W^{Q}\), as described below:
\begin{equation}
\begin{aligned}
    I\_embed = I\_Encoder(image)W^{I}, \\
    Q\_embed = T\_Encoder(question)W^{Q} \label{encode_input}
\end{aligned}
\end{equation}

The embeddings from the image and text encoders are concatenated to form a joint feature representation (\(F\_embed\)). This joint embedding is further transformed using a linear projection matrix \(W^{F}\) to integrate the two modalities:
\begin{equation}
\begin{aligned}
    F\_embed = Concat(I\_embed, Q\_embed)W^{F} \label{feature_concat}
\end{aligned}
\end{equation}

The joint feature embedding (\(F\_embed\)) is passed through a classifier to predict the output classes. This classifier applies a ReLU activation, defined as:
\begin{equation}
    ReLU(x) = \max(0, x)
\end{equation}
followed by a linear layer with weight matrix \(W^{CLS}\). The output of the classifier is computed as:
\begin{equation}
\begin{aligned}
    Classifier = ReLU(F\_embed)W^{CLS}
\end{aligned}
\end{equation}

This architecture defines projections as learnable linear layers, where the weight matrices are defined as follows: \(W^{I} \in \mathbb{R}^{d \times d\_{img}}\), \(W^{Q} \in \mathbb{R}^{d \times d\_{text}}\), and \(W^{F} \in \mathbb{R}^{2 \times d \times d}\). 

For this work, we use a hidden size of \(d = 512\), with the embedding size from the image encoder set to \(d_{img} = 768\) and the output embedding size from the text encoder set to \(d_{text} = 1024\). The classifier is implemented with a simple ReLU activation followed by a linear layer, with the weight matrix \(W^{CLS} \in \mathbb{R}^{d \times C}\), where \(C\) is the number of output classes.

\subsection{Textual Feature Augmentation}

\subsubsection{Paraphrase Pool.} 
Paraphrasing is our augmentation strategy to enhance the training dataset with existing questions. This approach allowed us to enrich the dataset with a wealth of linguistic variations, which were subsequently incorporated into the training dataset before model training. Each question in the training set is paraphrased by a pre-trained model tailored for this task. An mT5 \cite{xue2020mt5} model fine-tuned on a Vietnamese dataset translated from an English dataset is used. Particularly, a pool size of 10 paraphrases (\(\mathcal{P}\)) is considered in this method.

\subsubsection{Question Augmentation}
To enhance the representation of text input features, a specialized module is introduced within the network to craft augmented samples. Given a question $Q \in \mathcal{Q}$ and two paraphrased questions, $p_{i}$, randomly sampled from a pool $\mathcal{P}$, the following steps are applied:

First, the embedding vectors for the original question and the paraphrased questions are extracted using a transformer encoder. The embedding for the original question, $E^{O}$, is computed as:
\begin{equation}
\label{eq:original_embedding}
E^{O} = T\_Encoder(Q),
\end{equation}
where $T\_Encoder$ denotes the Transformer-based encoder used to produce text embeddings.

For each paraphrased question $p_{i}$, its embedding $E^{P}_{i}$ is computed and transformed using a weight matrix $W^{P}$ specific to paraphrased questions:
\begin{equation}
\label{eq:paraphrased_embedding}
E^{P}_{i} = T\_Encoder(p_{i})W^{P}, \quad i \in \{0, 1\},
\end{equation}

Next, the embedding of the original question, $E^{O}$, is linearly transformed using a weight matrix $W^{O}$:
\begin{equation}
\label{eq:transformed_original}
E^{O}_{transformed} = E^{O}W^{O},
\end{equation}

The transformed embedding $E$ is then calculated as the sum of the transformed original embedding and the paraphrased embeddings:
\begin{equation}
\label{eq:combined_embedding}
E = E^{O}_{transformed} + \sum_{i=0,1} E^{P}_{i}.
\end{equation}

To introduce non-linearity and improve representation power, a ReLU activation function is applied to $E$. A skip-connection is then added from the original question embedding to form the final augmented embedding, $Q_{augmented}$:
\begin{equation}
\label{eq:augmented_embedding}
Q_{augmented} = ReLU(EW^{out}) + E^{O}.
\end{equation}
where $W^{P} \in \mathbb{R}^{d \times d_{k}}$, $W^{O} \in \mathbb{R}^{d \times d_{k}}$ and  $W^{out} \in \mathbb{R}^{d_{k} \times d}$ are learnable parameters. For each of these we use $d_{k} = d_{text} / 2 = 512$.

Finally, to inject $Q_{augmented}$ as the embedding $Q_{embed}$, the weight matrix $W^{Q}$, previously defined in Equation~\ref{encode_input}, is applied:
\begin{equation}
\label{eq:final_embedding}
Q_{embed} = Q_{augmented}W^{Q}.
\end{equation}

This process ensures that the question embedding incorporates both the original and paraphrased information, thereby enriching the feature space and enhancing the model's ability to handle linguistic variability.

\subsubsection{Cross-training with origin and augmented question.}

To incorporate the proposed augmentation during training, two distinct processing branches are introduced within the text channel of the VQA model, following the text encoder. Each branch provides a unique input representation to the classifier. The first branch directly forwards the original output of the text encoder to the concatenation phase. In contrast, the second branch applies the question augmentation process, generating an augmented version of the question embedding to enhance the model's exposure to diverse textual representations. To control the augmentation process, which determines the extent to which the model learns on augmented embeddings, a random value \(x \sim \mathcal{U}(0, 1)\) is generated for each forward pass. If \(x < t_{thresh}\), the paraphrased version is used; otherwise, the original question is retained. This stochastic process enables a balanced exposure to both original and augmented data, fostering the model's ability to generalize across different textual representations. 

It is also noted that only the non-augmented branch is activated during inference, as there is no involvement of paraphrases in the test environment, in which the use of additional external knowledge is not permitted.

\subsection{Curriculum Learning} 
\begin{figure}[ht]
    \centering
    \includegraphics[width=\linewidth]{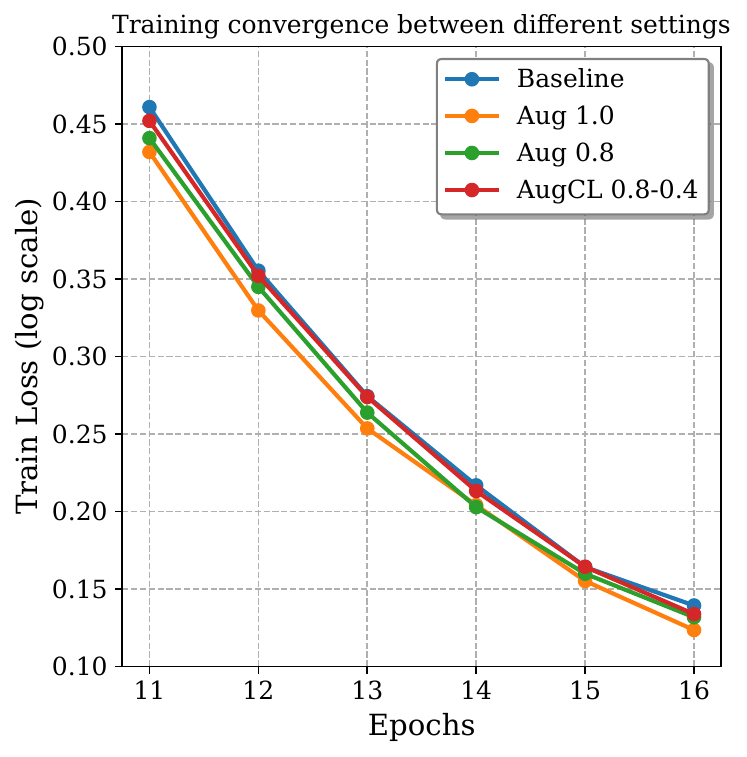}
    \caption{Visualization of training loss (log scale) of the model under different dataset configurations. Baseline uses no augmented samples, Aug 1.0 and Aug 0.8 use fixed thresholds of 1.0 and 0.8 for augmentation, and AugCL 0.8-0.4 applies curriculum learning with thresholds decreasing from 0.8 to 0.4. Training with paraphrased embeddings improves convergence compared to the baseline, with the selected epoch range (11–16) capturing critical optimization stages.}
    \label{fig:convergence}
\end{figure}

When analyzing the training performance of a VQA model on non-augmented and augmented datasets, moderate improvements in convergence were observed in cases involving enhanced embeddings (see Figure \ref{fig:convergence}). We hypothesize that embeddings fused with paraphrased sentences capture richer and more diverse patterns for the model to learn from, compared to embeddings directly generated by the text encoder from the original input. This observation inspired us to apply curriculum learning to the training scheme to expose the model to simple samples first and gradually transition to more complex ones. By doing so, the model builds a stable foundation for learning, resulting in more consistent training and improved convergence, while also leveraging the established success of curriculum learning in enhancing model generalization. Specifically, enhanced samples, which contain question embeddings combined with paraphrase information, are considered ``easy'' samples, while raw samples are regarded as ``hard'' samples.

Because the proportion of augmented and raw embeddings in the training dataset can be adjusted using a threshold value, as mentioned earlier, the traditional idea of progressively exposing the model to harder samples is slightly modified. Instead, the rate of exposure to hard samples is increased until the end of training. This is achieved by dynamically updating the threshold value based on the training epoch, ensuring a gradual shift in sample composition while maintaining a balance between easy and hard samples throughout the process.

Afterwards, a strategy is investigated for adjusting \(t_{\text{thresh}}\), the threshold that determines the proportion of augmented (easy) and raw (hard) samples during training. Inspired by prior research on progressive learning schedules, this strategy uses a simple linear decay formula to update \(t_{\text{thresh}}\) over the course of training. Specifically, \(t_{\text{max}}\) and \(t_{\text{min}}\) represent the initial and minimum values of \(t_{\text{thresh}}\), while \(T\) is the total number of epochs and \(e_{\text{cur}}\) is the current epoch. The formula for this update is as follows:

\begin{equation}
    t_{\text{thresh}} = \max\left(t_{\text{max}} - \frac{(t_{\text{max}} - t_{\text{min}})}{T} \cdot e_{\text{cur}}, \, t_{\text{min}}\right) 
\end{equation}

Using this formula, \(t_{\text{thresh}}\) decreases linearly from \(t_{\text{max}}\) to \(t_{\text{min}}\) as training progresses. This ensures a steady and predictable transition, gradually shifting the model's focus from simpler (augmented) data to more challenging (raw) samples at a uniform rate.

\section{Experiments}

\subsection{Experimental Settings}

\subsubsection{Datasets.}

We conduct experiments on two Vietnamese Visual Question Answering datasets: ViVQA \cite{tran-etal-2021-vivqa} and OpenViVQA \cite{openvivqa}. As VQA is still a relatively new task in the Vietnamese language domain, both datasets represent pioneering efforts to address the scarcity of Vietnamese VQA resources. These datasets consist of open-ended questions paired with images and are publicly available. The training set of each dataset is used to train the proposed method, and its performance is evaluated on the test-dev set. 

\begin{itemize} 
\item \textbf{ViVQA:} consists of 15,000 question-answer pairs derived from 10,328 images in the MS COCO dataset. The questions were translated into Vietnamese from the original English dataset using a semi-automatic translation process, ensuring linguistic diversity and alignment with Vietnamese semantics.
\item \textbf{OpenViVQA:} provides a richer structure with 11,000+ images and 37,000+ question-answer pairs, specifically designed for generative VQA tasks. It includes answers represented in natural language rather than fixed options, making it suitable for evaluating models that provide free-form responses. This dataset is publicly available as part of the VLSP 2023 - ViVRC shared task challenge.
\end{itemize}

\subsubsection{Evaluation Metrics.}
Given the differing nature of the datasets, evaluation metrics tailored to each are employed. For the ViVQA dataset, which is designed for classification-based VQA tasks, accuracy is used. Accuracy measures the proportion of correctly predicted answers to the total number of questions and is defined as:
\begin{equation}
\text{Accuracy} = \frac{\text{Number of Correct Answers}}{\text{Total Number of Questions}}
\end{equation}

In contrast, for OpenViVQA, which focuses on generative-based VQA tasks, CIDEr \cite{cider} is used for evaluation. This metric measures how closely a model-generated answer matches a set of reference answers by comparing shared \(n\)-grams and is calculated as:
\begin{equation}
\text{CIDEr} = \frac{1}{N} \sum_{i=1}^{N} \sum_{n=1}^{4} w_n \cdot \frac{g_n(c_i, r_i)}{\sqrt{g_n(c_i, c_i) \cdot g_n(r_i, r_i)}}
\end{equation}

In this formula, \(c_i\) denotes the prediction from the model, while \(r_i\) represents the set of reference answers. The weight \(w_n\) indicates the significance assigned to \(n\)-grams of varying lengths, and \(g_n\) measures the match of \(n\)-grams between the predicted answer and the reference answers. By analyzing the overlap of key phrases, CIDEr evaluates how closely the hypothesized response aligns with the references.

These metrics are used consistently throughout this paper to benchmark the performance of our proposed methods in both classification-based and generative-based VQA tasks.


\subsubsection{Implementation Details.}
Throughout the experiment, all models are trained for 40 epochs using the AdamW \cite{loshchilov2018decoupled} optimizer with a learning rate of 1e-5 and a batch size of 16. Early stopping with a patience value of 5 is applied to ensure optimal convergence. To improve training efficiency, mixed precision training \cite{micikevicius2017mixed} is employed.

To better handle the linguistic features of the Vietnamese language, the text encoder uses Vietnamese pre-trained models. The transformer-based Vietnamese text encoders BARTpho \cite{luong2021bartpho} and PhoBERT \cite{nguyen-tuan-nguyen-2020-phobert} are the two considered in this work. For image feature extraction, ResNet18 \cite{he2016deep} is used as a representative CNN-based encoder for capturing spatial image features, while BEiTv2 \cite{peng2022beit} serves as a transformer-based model for extracting global image features. Since the vanilla VQA model is framed as a classification approach, this formulation is maintained even for the OpenViVQA dataset, which was originally designed for generative models. This ensures consistency in evaluation and enables us to directly compare results across datasets. For each experiment, models are trained with 5 distinct random seeds, and the mean as well as standard deviation are reported. All experiments are conducted on a single NVIDIA RTX A5000 GPU.

\subsection{Experimental Results}

We present the results of experiments conducted to evaluate the effectiveness of the proposed training pipeline on a vanilla VQA model using the OpenViVQA dataset. The text and image encoders are replaced with different backbone combinations. Each backbone setup is treated as a distinct model, and the proposed method is systematically applied to assess its impact on the baseline. In specific:

\begin{itemize} \item \textbf{B:} The baseline method is trained in a standard manner using original representations from the encoders.
\item \textbf{B + Aug:} The baseline model trained with curriculum learning, where a dynamically adjusted threshold determines the choice between raw and augmented representations during training. The scheme is configured to be started at the threshold of $0.8$, then linearly decrease to $0.4$ during training.
\end{itemize}

The results in Table~\ref{tab:results} indicate that the proposed method (B + Aug) consistently outperforms the baseline (B) across all backbone combinations. The BARTpho-word encoder achieves notable improvements, in particular with ResNet18, which records the highest mean CIDEr score. Furthermore, the proposed approach demonstrates greater stability, evident from reduced standard deviation in several configurations. These results highlight the effectiveness of the training pipeline in enhancing the model's robustness and overall performance.

\begin{table*}[!ht]
\centering
\caption{Evaluation results of the baseline (B) and proposed method (B + Aug) on the OpenViVQA dataset using different text and image encoders. Results are measured using CIDEr. Mean and standard deviation are reported over 5 random seeds. Bold values indicate the best performance for each configuration.}
\label{tab:results}
\begin{tabular}{l|l|cc|cc}
\hline
\textbf{Text Encoder} & \textbf{Image Encoder} & \multicolumn{2}{c|}{\textbf{B}} & \multicolumn{2}{c}{\textbf{B + Aug}} \\ 
                      &                        & \textbf{Mean}    & \textbf{Std}       & \textbf{Mean}    & \textbf{Std}        \\ \hline
\multirow{2}{*}{BARTpho-word} 
                      & ResNet18               & 0.5656          & 0.0415            & \textbf{0.5874}  & 0.0301              \\ 
                      & BEiTv2                 & 0.4790          & 0.0609            & \textbf{0.5553}  & 0.0174              \\ \hline
\multirow{2}{*}{PhoBERT-base} 
                      & ResNet18               & 0.3371          & 0.0560            & \textbf{0.3558}  & 0.1143              \\ 
                      & BEiTv2                 & 0.3337          & 0.0627            & \textbf{0.3420}  & 0.0757              \\ \hline
PhoBERT-large         & ResNet18               & 0.1660          & 0.0974            & \textbf{0.2301}  & 0.1475              \\ \hline
\end{tabular}
\end{table*}

\subsection{Ablation Study}
Our proposed method involves several hyperparameters that significantly influence the performance of the model learning process. To evaluate their impact, a series of experiments on the ViVQA dataset is conducted using the BARTpho+ResNet backbone as the model to ensure consistency and focus on analyzing the effects of the hyperparameters. The influence of these adjustments on model performance is examined, and the results are presented in detail.

\subsubsection{Fixed Threshold ($t_\text{thresh}$).} 

To assess how learning on a mix of raw and augmented embeddings controlled by a fixed threshold benefits the model, experiments are conducted with fixed threshold values for the ratio of raw to augmented samples, denoted as \(t_\text{thresh}\). These values range from 0 to 1 in increments of 0.2. A threshold ratio closer to 1 represents a higher proportion of augmented samples, while a ratio closer to 0 represents a higher proportion of raw samples. This exploration serves not only to determine the optimal balance for effective learning but also as a reference for comparing with a dynamic thresholding approach.

\begin{table}[ht]
\centering
\caption{Experimental results for evaluating the impact of the thresholding ratio.}
\label{tab:thresholding_results}
\begin{tabular}{lc}
\hline
\textbf{Threshold Ratio ($t_\text{thresh}$)} & \textbf{Accuracy} \\ \hline
0.0 (baseline)                   & 0.5432 ± 0.0074           \\
0.2                              & 0.5446 ± 0.0028           \\
0.4                              & 0.5547 ± 0.0060           \\
0.6                              & 0.5525 ± 0.0064           \\
0.8                             & \textbf{0.5554 ± 0.0043}  \\
1.0                              & 0.5283 ± 0.0085           \\ 
\hline
\end{tabular}
\end{table}

The results summarized in Table~\ref{tab:thresholding_results}, show that the accuracy is influenced by the choice of \(t_\text{thresh}\). Using only augmented samples (\(t_\text{thresh} = 1\)) results in the lowest accuracy, which is expected since the distribution of embeddings from the training data differs significantly from the embeddings of the test data. Conversely, using only raw samples (\(t_\text{thresh} = 0\)) achieves slightly better performance (\(0.5432 \pm 0.0074\)), but the best accuracy (\textbf{0.5554 ± 0.0043}) is achieved with \(t_\text{thresh} = 0.8\). This suggests that a majority of augmented samples, combined with a smaller proportion of raw samples, provides an optimal balance for effective learning.

These findings highlight that a carefully chosen threshold ratio is critical for maximizing the benefits of interleaving raw and augmented samples in training. 

\subsubsection{Number of Paraphrases in Augmented Embedding.} 
As stated in our motivation, the use of paraphrases helps the model learn more easily by enriching the original representation of the input question. We argue that adding more paraphrases results in better representations as more information is incorporated, potentially improving the model learning process. To evaluate this, the number of paraphrases sampled from the paraphrase pool is varied, with configurations of 0 (baseline), 1, 2, and 3 paraphrases per question tested with the proposed method.

\begin{table}[ht]
\centering
\caption{Experimental results for varying the number of paraphrases under the proposed curriculum learning framework.}
\label{tab:paraphrase_results}
\begin{tabular}{lc}
\hline
\textbf{Number of Paraphrases} & \textbf{Accuracy} \\ \hline
0 (baseline)                   & 0.5432 ± 0.0074    \\
1                              & 0.5531 ± 0.0021     \\
2                              & 0.5538 ± 0.0038      \\
3                              & \textbf{0.5547 ± 0.0039}       \\ \hline
\end{tabular}
\end{table}

Concretely, it is confirmed in Table~\ref{tab:paraphrase_results} that the use of paraphrases enhances performance compared to the baseline, with accuracy improving as the number of paraphrases increases. This trend highlights the effectiveness of leveraging multiple paraphrases to enrich the input representation and improve the model’s learning process under the curriculum learning framework. However, increasing the number of paraphrases also leads to a higher demand for GPU memory (VRAM), making the training process more resource-intensive. To balance performance improvement with computational efficiency, \(n = 2\) is chosen for the proposed method. This choice provides a practical trade-off, ensuring notable accuracy gains while keeping the resource requirements manageable.

\subsubsection{Epoch Update Strategy in Curriculum Learning.}

In our setting, the difficulty level of dataset is dynamically regulated by a threshold (\(t_\text{thresh}\)) that adjusts over epochs, governing the presence of easy (augmented) and hard (raw) samples during training to ensure the model incrementally builds robust representations. Initially, the model focuses on simpler samples to stabilize learning and gradually incorporates more challenging ones as training progresses. To evaluate the effectiveness of different update strategies, two approaches for adjusting \(t_\text{thresh}\) are considered: Linear Decay and Cosine Annealing. In both cases, the threshold is decreased from a specified maximum value (\(t_\text{thresh, max}\)) to a minimum value (\(t_\text{thresh, min}\)) over the training time (see Table~\ref{tab:update_strategy_results}).

\begin{table}[ht]
\centering
\caption{Experimental results for different epoch update strategies in curriculum learning. The highest accuracy for each decay strategy is marked in bold.}
\label{tab:update_strategy_results}
\begin{tabular}{lcc}
\hline
\textbf{Decay Strategy}    & \textbf{$t_\text{thresh, max/min}$} & \textbf{Accuracy} \\ \hline
\multirow{8}{*}{Linear} 
                          & 1.0 / 0.8        & \textbf{0.5583 ± 0.0068} \\
                          & 1.0 / 0.6        & 0.5565 ± 0.0052 \\
                          & 1.0 / 0.4        & 0.5566 ± 0.0018 \\
                          & 1.0 / 0.2        & 0.5524 ± 0.0053 \\
                          & 1.0 / 0.0        & 0.5518 ± 0.0061 \\
                          & 0.8 / 0.6        & 0.5553 ± 0.0062 \\
                          & 0.8 / 0.4        & 0.5560 ± 0.0062 \\
                          & 0.8 / 0.2        & 0.5523 ± 0.0046 \\
                          & 0.8 / 0.0        & 0.5531 ± 0.0024 \\ \hline
\multirow{5}{*}{Cosine Annealing} 
                          & 1.0 / 0.8        & 0.5489 ± 0.0061 \\
                          & 1.0 / 0.6        & \textbf{0.5547 ± 0.0047} \\
                          & 1.0 / 0.4        & 0.5525 ± 0.0045 \\
                          & 1.0 / 0.2        & 0.5510 ± 0.0026 \\
                          & 1.0 / 0.0        & 0.5541 ± 0.0063 \\
\hline
\end{tabular}
\end{table}

The results are summarized in Table~\ref{tab:update_strategy_results}. For the Linear Decay strategy, the best performance is achieved when \(t_\text{thresh}\) decreases linearly from 1.0 to 0.8, yielding an accuracy of \(0.5583 \pm 0.0068\). However, reducing \(t_\text{thresh}\) further to lower minimum values (e.g., 0.2, 0.0) leads to a consistent drop in performance, indicating that excessively emphasizing harder samples can destabilize learning.

In contrast, the Cosine Annealing strategy exhibits a different trend. While it does not achieve the best overall performance (in the context of the experiment of Table~\ref{tab:thresholding_results}), it shows competitive results, particularly when \(t_\text{thresh}\) is reduced from 1.0 to 0.6, achieving an accuracy of \(0.5547 \pm 0.0047\). Interestingly, Cosine Annealing is slightly less sensitive to smaller minimum threshold values compared to Linear Decay, with its performance remaining relatively stable for \(t_\text{thresh, min} = 0.0\) (\(0.5541 \pm 0.0063\)). These findings highlight that while both strategies are effective, Linear Decay with a moderate reduction in \(t_\text{thresh}\) provides the best balance between learning stability and exposure to harder samples. 

\begin{figure}[ht]
    \centering
    \includegraphics[width=\linewidth]{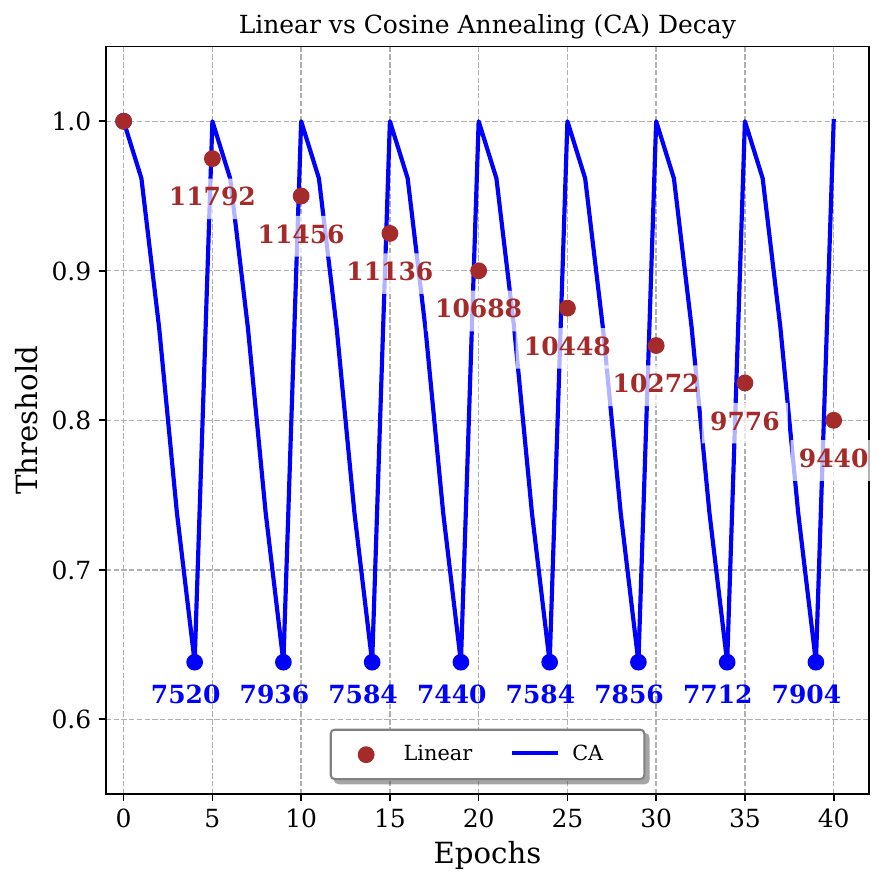}
    \caption{Comparison of Linear and Cosine Annealing (CA) decay strategies for updating \(t_\text{thresh}\) over training epochs. The numbers on each line indicate the number of augmented samples used for training at that epoch. In the experiments in Table~\ref{tab:update_strategy_results}, Linear decay achieves the best performance when \(t_\text{thresh}\) decreases linearly from 1.0 to 0.8. For Cosine Annealing decay, the best performance is achieved when \(t_\text{thresh}\) is reduced to 0.6.}
    \label{fig:linear_vs_cosine_annealing_decay}
\end{figure}

A more comprehensive comparison of both techniques is given in Figure~\ref{fig:linear_vs_cosine_annealing_decay}, highlighting the behavior of each method. Linear decay updates the threshold value only after the completion of an epoch, whereas Cosine Annealing adjusts it at every iteration.

\subsubsection{Performance on ViVQA Across Different Backbones.}
In this part, the overall performance of the proposed method is evaluated on the ViVQA dataset using various backbone settings. This evaluation aims to analyze how different configurations of text and image encoders impact the effectiveness of the proposed training pipeline.

\begin{table}[ht]
\centering
\caption{Evaluation results of the baseline (B) and proposed method (B + Aug) on the ViVQA dataset using different text encoders and ResNet18 as the image encoder. Results are measured using Accuracy. Bold values indicate the best performance for each configuration.}
\label{tab:vivqa_results}
\begin{tabular}{l|cc}
\hline
\textbf{Text Encoder} & \textbf{B}           & \textbf{B + Aug}         \\ \hline
BARTpho-word          & 0.5432 ± 0.0074     & \textbf{0.5538 ± 0.0038} \\ 
PhoBERT-base          & \textbf{0.5453 ± 0.0047} & 0.5373 ± 0.0029         \\ 
PhoBERT-large         & 0.5435 ± 0.0204     & \textbf{0.5486 ± 0.0069} \\ \hline
\end{tabular}
\end{table}

The Table~\ref{tab:vivqa_results} shows mixed outcomes across text encoders, with only small improvements observed in some cases. For BARTpho-word and PhoBERT-large, the proposed method slightly outperforms the baseline, while for PhoBERT-base, the baseline achieves marginally better performance. We argue that this performance variation is due to the limited size of the ViVQA dataset and the inclusion of low-quality auto-translated samples, which might introduce confusion into the paraphrases and result in suboptimal augmentation quality. These factors likely hinder the ability of our proposed method to fully demonstrate its potential.

\section{Conclusion and Future Work}
In this paper, we proposed a novel training pipeline for Vietnamese VQA models, incorporating curriculum learning to train on a combination of raw and augmented textual features. By progressively transitioning from a greater to a smaller reliance on augmented samples, the model learns to handle varying levels of complexity during training. These augmented samples capture the essence of paraphrased variations of the input question, enabling the model to improve its generalization and versatility in handling the inherent variability of the Vietnamese language. Extensive experiments demonstrate that this method enhances the performance of baseline models, providing an effective approach to advancing VQA in underrepresented languages. This strategy is considered promising for Vietnamese VQA and lays the foundation for future advancements, with potential for adaptation to other low-resource languages.

Future development based on this method could focus on several promising directions. First, developing a learnable approach to optimize hyperparameters, rather than relying on manually determined constants, could streamline the training process and improve overall performance. Second, to fully exploit the potential of image features, future work could explore augmentation techniques for the image channel and design efficient strategies to train with both augmented image and text modalities. Third, extending the proposed framework to additional languages would validate its effectiveness across diverse linguistic contexts and contribute to the development of VQA systems for low-resource settings. Finally, further investigation into curriculum learning strategies, such as adaptive curriculum schedules or dynamically adjusting difficulty levels based on model performance, could unlock additional gains and make the training process even more effective.

\section*{Acknowledgements}

This research was fully supported by AI VIETNAM.


\bibliography{aaai25}

\end{document}